\documentclass[pre,superscriptaddress,twocolumn,showkeys,showpacs]{revtex4}
\usepackage{epsfig}
\usepackage{latexsym}
\usepackage{amsmath}
\begin {document}
\title {Language structure in the $n-$object naming game}
\author{Adam Lipowski}
\affiliation{Faculty of Physics, Adam Mickiewicz University, 61-614
Pozna\'{n}, Poland}
\author{Dorota Lipowska}
\affiliation{Institute of Linguistics, Adam Mickiewicz University,
60-371 Pozna\'{n}, Poland}
\begin {abstract}
We examine a naming game with two agents trying to establish a
common vocabulary for $n$ objects. Such efforts lead to the
emergence of language that allows for an efficient communication
and exhibits some degree of homonymy and synonymy. Although
homonymy reduces the communication efficiency, it seems to be a
dynamical trap that persists for a long, and perhaps indefinite,
time. On the other hand, synonymy does not reduce the efficiency
of communication, but appears to be only a transient feature of
the language. Thus, in our model the role of synonymy decreases and in the long-time limit
it becomes negligible. A similar rareness of synonymy is observed in present
natural languages. The role of noise, that distorts the
communicated words, is also examined. Although, in general, the
noise reduces the communication efficiency, it also regroups the
words so that they are more evenly distributed within the
available "verbal" space.
\end{abstract}
\maketitle
\section{Introduction}
Computational modelling is becoming more and more important tool
to study langauge
evolution~\cite{DEBOER,NOWAK,NOWAK2002,PAULO2007,STAUFFER2008}.
The central assumption of such an approach is that language is a
complex adaptive system that emerges from local interactions
between its users, and evolves and complexifies according to
biological-like principles of evolution and
self-organization~\cite{BRIGHTON2005,CHRISTIANSEN,STEELS2002}.
This is by no means the only possibility since a number of
researches claims that language does not have the adaptive values
and is merely a byproduct of having a large and complex brain or
of some other skills~\cite{CHOMSKY1972,GOULD}. Recently, however,
adaptationists got a strong support from Pinker and Bloom, who in
their influential paper~\cite{PINKER1990} argued that linguistic
abilities require complex and costly adaptations (e.g., large
brain, longer infancy period, descended larynx) and the language
origin can be explained only by means of natural selection theory.

Since language was invented only in one lineage, and is therefore
unique to human species, its appearance has the same status as the
origin of genetic code or the eukaryotic cell. The emergence of
language was thus listed as one of the major transition in the
evolution of life on Earth~\cite{MSMITH1997} and it is certainly
interesting to ask which factor is responsible for it. Some claims
were made that most likely it was the combination of selective
evolutionary pressure and unique context that lead to the
emergence of human language~\cite{SZAMADO}.

Language has also lead to the novel inheritance
system~\cite{JABLONKA} and opened up the possibility for
cumulative cultural evolution and creation of complex
society~\cite{MAYNARD2003} with collaboration of large non-kin
groups~\cite{FEHR2002}. While our willingness to share information
with relatives is rather easy to reconcile with darwinian
evolution (kin-selection~\cite{HAMILTON1964}), linguistic
interactions with non-kin individuals are harder to understand.
Indeed, since speaking is costly (it takes time, energy and
sometimes might expose a speaker to the predators), and listening
is not, such a situation seems to favour selfish individuals that
would only listen but would not speak. Moreover, in the case of
the conflict of interests the emerging communication system would
be prone to misinformation or lying. A possible resolution of
these problems is based on reciprocal altruism~\cite{TRIVERS}.
However, there is a growing evidence that cooperation and
altruistic behaviour between humans are very complex and typically
cannot be explained using standard reciprocal altruism
arguments~\cite{FEHR2002}.

As an alternative explanation Dessalles\cite{DESSALLES} suggests
that honest information is given freely because it is profitable -
it is a way of competing for status within a group. In this
context, an interesting computer simulations were made by
Hurford\cite{HURFORD}. He considered agents engaged in
communicative tasks (one speaker and one hearer) and their
abilities evolved with the genetic algorithm that was set to
prefer either communicative or interpretative success. Only in the
former case the emerging language was similar to natural languages
where synonymy was rare and homonymy tolerated. When interpretative
success was used as the basis of selection then the converse
situation (unknown in natural language) arose: homonymy was rare
and synonymy tolerated.

Indeed, synonymy in the pure variety is rare. Usually, it can be
found in two languages being in contact (\emph{napkin/serviette}),
handy abbreviations (\emph{bicycle/bike}) or some specialized
euphemistic domains related e.g., with sex
(\emph{fuck/shag/}\ldots), death (\emph{croak/expire/}\ldots) or
bodily functions (\emph{shit/crap/}\ldots). Linguists proposed
various explanations of the human avoidance of synonymy. Clark
attributes it to a presumably inborn tendency of humans to seek
and create new meanings, rather than accept one meaning for
several different forms~\cite{CLARK}. Markman notes that children
have a tendency to assume that no two words may overlap in
meaning~\cite{MARKMAN}. A similar point of view is expressed in
Wexler's Uniqueness Principle which prevents the child from
internalizing more than one form per meaning~\cite{WEXLER}. On the
other hand, homonymy seems to be more common in natural languages.
One can easily think of many words having multiple and unrelated
meanings (e.g., \emph{abstract, compound, second, present}). At
first sight one can consider this as surprising since synonymy
does not diminish communicative efficiency but homonymy in
principle does. Let us also notice that computer languages quite
often accept synonymy (e.g., aliases in command systems) but
typically do not handle homonymy.

In our opinion an apparent asymmetry between rare synonymy and
relatively common homonymy is an important and generic feature of
natural languages and might be used as a test of various
computational models of language development. In the present paper
we examine a version of the Steels naming game
model~\cite{STEELS1995} where two agents exchange information
concerning a certain number of facts/objects from their reality
 and try to establish a common vocabulary. The emerging language
features some degree of homonymy and synonymy. Although homonymy
diminishes the communicative efficiency it turns out to be a
persistent feature of the language. On the other hand synonymy is
only a transient feature of the language and its frequency of
appearance diminishes over time. The asymmetry between homonymy
and synonymy can be thus understood within a rather simple
naming-game setup, without revoking evolutionary arguments that
speaker more than hearer benefits from the
conversation~\cite{HURFORD}.
Let us also notice that stable homonymy and transient synonymy was also reported by Puglisi \emph{et al.} in a model of formation of categories~\cite{PUGLISI}.
We also examine the role of noise that might distort communicated words. Our results show that noise plays (or played) an important role and could affect the distribution of words in a "verbal" space.
\section{Model}
More than a decade ago Steels proposed the naming game model, that
quickly became one of the basic models of the emergence of
linguistic coherence~\cite{STEELS1995}. In this model we have a
group of agents that communicate with each other trying to
establish a common vocabulary on a certain number of objects.
Typically, after some time, they reach a state of linguistic
coherence where they to large extent (or even perfectly)
understand each other. In the original formulation the naming game
model describes cultural transmission within a single generation
of agents. Evolutionary versions with mutations and selections of
agents taking place were also
studied~\cite{LIPLIP2008,LIPLIP2008A}. In most works on the naming
game model only a simple structure of the emerging language is allowed and homonymy is very often excluded~\cite{BARONCHELLI2006,ASTA2006, BARONIJMP}. Such approaches effectively can be regarded as if agents would talk on a single object~\cite{BARONIJMP}. Although it drastically simplifies the language structure such an approach allows to consider many agents and thus to take into account some elements of the social structure~\cite{ASTA2006}. But such works constitute only one end point of the computational dilemma: many agents of simple-architecture versus few but with complex-architecture. At the other side we have models of few agents but able to develop language of much larger complexity. To examine linguistic structures like homonyms or synonyms, one has to consider an $n$-object
version of the naming game model. Some results on $n$-object
naming game model have been already
reported~\cite{LENAERTS,STEELSINTYRE}. Let us also notice that the
main emphasis in the naming game model is on the cultural
(single-generation) transmission of language. An alternative
approach to the language evolution were inter-generational
interactions play an important role is called Iterated Learning
Model and was used in various contexts~\cite{KIRBY2001}.

Our model is a two-agent version of the naming game. It is assumed
that agents are embodied in a shared environment and communicate
on a certain number of facts/objects from this environment. Agents
in turns take the role of speaker and hearer. Speaker selects an
object from the environment. Then, using its form-meaning
relations, speaker selects a word that is assigned to the object.
The word is communicated to hearer, that uses its own meaning-form
relations to guess the communicated meaning. We also assume that
after such a communication attempt there is a possibility to check
(e.g., by pointing at the object) whether hearer guessed the
communicated meaning correctly. Established in such a way success
or failure modifies the structure of meaning-form relations of
agents to facilitate future communication attempts.

Both agents refer to the common set of $n$ objects and with each
object each agent relates the corresponding inventory (inventories
are numbered from 1 to $n$). Each inventory stores up to $l$ words
that are used to describe the corresponding object. With each word
in a repository the weight $w$ is associated that controls the
stochastic process of selecting a communicated word (speaker) and
decoding the meaning (hearer). The idea of assigning weights to words was already used in some
naming game models~\cite{STEELSINTYRE}. For computational purposes the
words are represented by integer numbers from 1 to $r$ but more
natural representations using strings of letters are also
possible. The parameter $r$ can be thus interpreted as
corresponding to the capacity of the "verbal space".
More detailed rules of our model are specified below:\\
\begin{itemize}
\item Speaker randomly selects an object. From the inventory that
corresponds to the selected object speaker selects the
communicated word $x_c$. The word is selected taking into account
the weights corresponding to each word in this inventory. We used
the method of roulette selection.
 \item Hearer tries to guess the meaning of the communicated word and decodes it.
To do that, hearer first calculates measures of similarity
$s^k(x_c)$ of the communicated word with $k$-th inventory
($k=1,2,\ldots,n$). The measures $s^k(x_c)$, that are calculated
using the following formula
\begin{equation}
s^k(x_c)=\frac{1}{\sum_i w_i} \sum_i
\frac{w_i}{\epsilon+|x_i-x_c|} \ \ k=1,2,\ldots,n, \label{eq1}
\end{equation}
 are then used to select the inventory that
fits the communicated word (roulette selection again but with
$s^k(x_c)$ as a weight of an inventory). In Eq.~(\ref{eq1}) $x_i$
and $w_i$ are the $i$-th word and its weight, respectively, and
the summation is over all elements of the $k$-th inventory
(numerated with $i$). The closer $x_i$ to the $x_c$ is, the larger
its contributions to the similarity measure $s^k(x_c)$ are. The
role of $\epsilon$ in Eq.~(\ref{eq1}) is to keep $s^k(x_c)$ finite
even when the communicated word is the same as one of the words in
the $k$-th inventory. Having calculated $s^k(x_c)$ for all
inventories, hearer uses the roulette selection to choose the
inventory that fits the communicated word. Since in our
calculations $\epsilon$ takes rather small values, inventories
that contain a communicated word (or words that are very close to
it) get large similarity measures and have larger probabilities of
being selected.
 \begin{itemize}
 \item  When the inventory selected by the hearer has the same number as
 that selected by speaker, we consider this as a communicative
 success. In such a case both agents increase the weights associated with
 the communicated word  by one. If in the hearer inventory
 there is no such a word (but it still has decoded the meaning
 correctly) we add the communicated word to this inventory with unit weight
 (if the inventory contains already $l$ elements we first remove the
 word with the smallest weight).
 \item When the inventory selected by the hearer has a different number than
 that selected by the speaker we consider this as a communicative
 failure. In such a case speaker decreases the weight associated with
 the communicated word  by one. Hearer inspects its inventory that
 has the same number as that selected by the speaker.
 If it contains the communicated word, its weight is increased.
 Otherwise, hearer adds the word to this inventory with unit
 weight.
\end{itemize}

Our simulations show that the model is relatively robust and small
changes of its rules or of values of parameters do not change much
the behaviour of the model. In particular, similar results are
obtained when an increase or decrease of weights in the case of
success or failure is done either with a fixed or weight-dependent
amount (e.g.  the larger the weight, the smaller the increase).
Let us also notice that the increase or decrease of weight in the
case of success or failure, respectively, resembles the
reinforcement learning approach and some naming game models with a
similar dynamics have been already examined~\cite{LENAERTS}.
 \item In some of our simulations we have examined the effect
 of noise that distorts the communicated word.
 More precisely, we assume that with the probability $p$ the communicated word chosen by
 speaker becomes
 \begin{equation}
 x\rightarrow x+\eta
 \label{noise}
 \end{equation}
 where $\eta$ is a random integer number uniformly drawn from the
 interval $<-a,a>$ and $a$ is the amplitude of noise (with the probability $1-p$
 the communicated word does not change). If $x$ calculated using
 Eq.~(\ref{noise}) happens to be outside the range $<1,r>$, a different
 instance of $\eta$ is generated.
\end{itemize}

\begin{figure}
\vspace{2cm} \centerline{ \epsfxsize=9cm \epsfbox{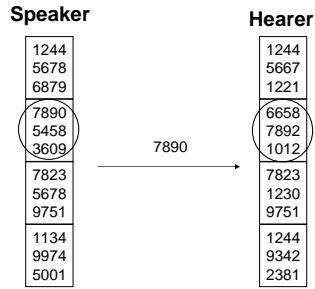} }
\vspace{-4cm} \caption{A communication attempt with $n=4$ objects.
Speaker selected the second object and one of the words that are
associated with this object. The selected word is communicated to
the hearer that then decodes its meaning. Since the decoded object
is the same as that chosen by the speaker, the above example is
considered as a success. The selection of the communicated word
and its decoding are stochastic in nature (see the main text) and
controlled by weights $w$ associated with each word.} \label{fig1}
\end{figure}

An example that illustrates the above rules is shown in
Fig.~\ref{fig1}. Table~\ref{tabela} collects all parameters of the model.

\begin{table}
\begin{tabular}{|c|c|}
\hline
 \textbf{param.} & \textbf{description (values used in simulations)} \\
\hline
 n & number of objects ($100\leq n\leq 10^3$)\\
\hline
 l & memory size - maximum number of words \\
   & corresponding to an object ($5\leq l\leq 20$)\\
\hline
 r & words - positive integer numbers not greater than r \\
   & ($500\leq r\leq 10^4$)  \\
 \hline
 $\epsilon$ & Ensures that similarity measure in Eq.~(\ref{eq1})\\
            & is finite (10$^{-5}\leq \epsilon\leq 10^{-1}$)\\
\hline
 $p,\ a$ & parameters describing noise\\
             & (see Eq.~\ref{noise}) ($0\leq p \leq 0.05,\ 0\leq\ a \leq 10$)\\
\hline
\end{tabular}
\caption{Parameters of the model and ranges of values used in the
simulations.\label{tabela}}
\end{table}
\section{Numerical Calculations}\
To start the simulations an initial configuration is needed. We
assume that at the beginning each agent has in each inventory a
single word (randomly selected from the interval $<1,r>$) with
unit weight. To examine the behaviour of the model we measured
various quantities that in some cases were averaged over certain
time intervals or over independent runs.( We define the unit of
time as corresponding to $2n$ communication attempts.) Of
particular interest is the communicative success rate of an agent,
that is defined as a fraction of successful communication
attempts. Some other quantities that allow us to analyze in more
details the structure of the emerging language and of the
communication process will be specified later.
\subsection{Basic properties}
\begin{figure}[!ht]
\vspace{4.5cm} \centerline{ \hspace{4cm}\epsfxsize=12cm
\epsfbox{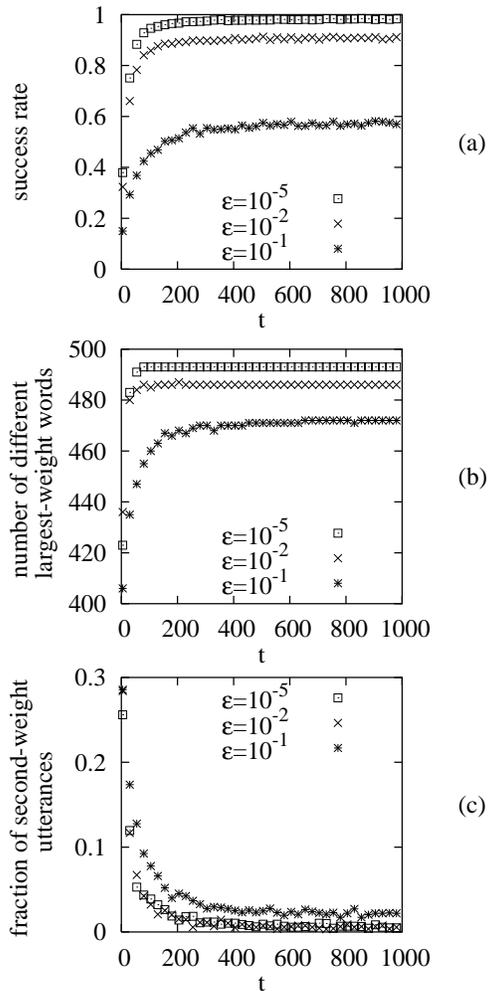} } \vspace{0.6cm} \caption{The time evolution
of (a) the success rate;  (b) the number of different
largest-weight words; and (c) the fraction of
second-largest-weight utterances. Calculations were made for
$n=500$, $l=10$ and $r=10^3$.} \label{epsilon}
\end{figure}

Simulations show that typically the agents correlate their
inventories so that their communication maintains a rather large
success rate (Fig~\ref{epsilon}a). Of our further interest will be
words that in a given inventory have the largest weight. Since
some of them might be the same for different inventories, we
calculated the number of different largest-weight words in the
resulting language. It turned out that this number is close to the
the number of objects $n$ (Fig~\ref{epsilon}b) and most of the
communication attempts use these largest-weight words
(Fig~\ref{epsilon}c). It means that in majority of cases
communication between agents proceeds as follows:  Speaker
selects an object and the largest-weight word from the inventory
corresponding to this object becomes the communicated word. For
small $\epsilon$ the similarity measure, as calculated from
Eq.(\ref{eq1}), is large only for the inventory that contains the
communicated word (provided that the weight of this word is not
very small). Usually, it happens to be the inventory corresponding
to the same object as selected by speaker and thus such an attempt
is successful. For larger $\epsilon\ (\sim 0.1)$ the communication
between agents deteriorates and both the success rate and the
number of different largest-weight words diminish.

There are two factors that contribute to the communication failure. First, it is the finite number of $\epsilon$ that implies that similarity measure (\ref{eq1}) for different words is positive and thus selection of the inventory made by the hearer might lead to communication failure. Second, homonyms, that as we shall see might appear in our model, also might lead to the communication failure. Their role is discussed in detail in the next subsection.

\begin{figure}[!ht]
\vspace{0.5cm} \centerline{ \hspace{0cm}\epsfxsize=9cm
\epsfbox{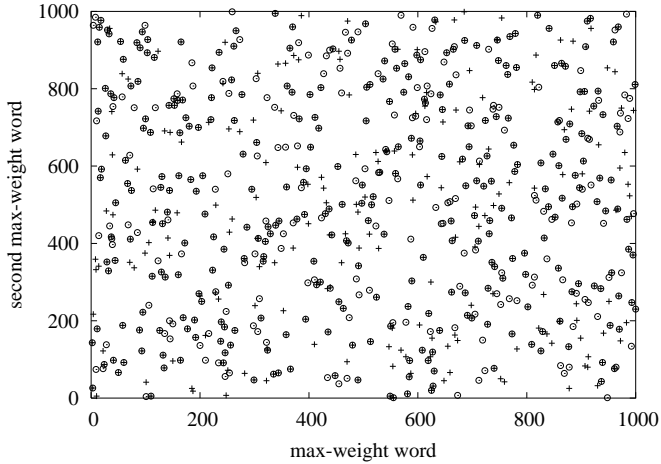} } \vspace{0cm} \caption{The distribution of
largest-weight words and the largest-weight words after
simulations of time $t=10^3$. Calculations were made for $n=500$,
$r=10^3$, $\epsilon=10^{-5}$, and and $l=10$. Different plotting symbols (circles,
crosses) correspond to different agents. Quite often both agents
have in some inventories (that usually corresponds to the same
object) the same largest- and second-largest-weight words and in
such a case the plotted symbols overlap.} \label{config0}
\end{figure}

In Fig.~\ref{config0} we present the distribution of largest- and
second-largest-weight words that is established after a
sufficiently long transient. Relatively uniform distribution
indicates that these words are uncorrelated. Since some of the
second-largest weight words, as discussed below, might be
considered as synonyms (of the largest-weight words), the lack of
correlations agrees with the observation that synonyms in natural
languages are not similar to each other.
\subsection{Homonymy and synonymy}
\begin{figure}
\vspace{2.5cm} \centerline{ \epsfxsize=9cm \epsfbox{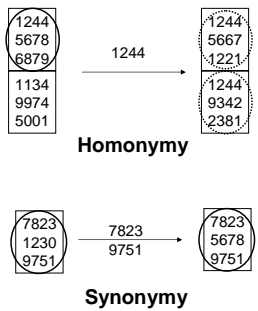}
} \vspace{-4cm} \caption{Homonymy occurs when a word (1244) can be
associated with more then one object. Synonymy occurs when an
object is associated with more than one word.} \label{homo-syno}
\end{figure}

Since agents communicate on more than one object the resulting
language might contain homonyms and synonyms. Homonymy appears
when a word can be associated with more than one objects and
synonymy when an object can be associated with several words.
However, the rules of our model contain probabilistic factors and
so the definition of homonymy and synonymy must take this fact
into account. We define homonymy as a word that with \emph{a
relatively large} probability can be associated with several
objects. Typically such a situation occurs when a word uttered by
the speaker appears in more than one inventories of the hearer as
the largest-weight word. Consequently, the number of different
largest-weight words is a measure of homonymy of the language (the
smaller this number is, the more frequent the homonymy is).
Analogously, synonymy most often occurs when speaker and hearer in
their inventories corresponding to a certain object have the same
largest- and second-largest-weight words. In such a case, no
matter which of them is selected for communication, it is quite
probable that the meaning will be guessed correctly. Examples of
such situations are shown in Fig.~\ref{homo-syno}.

\begin{figure}
\vspace{0.0cm} \centerline{ \epsfxsize=9cm
\epsfbox{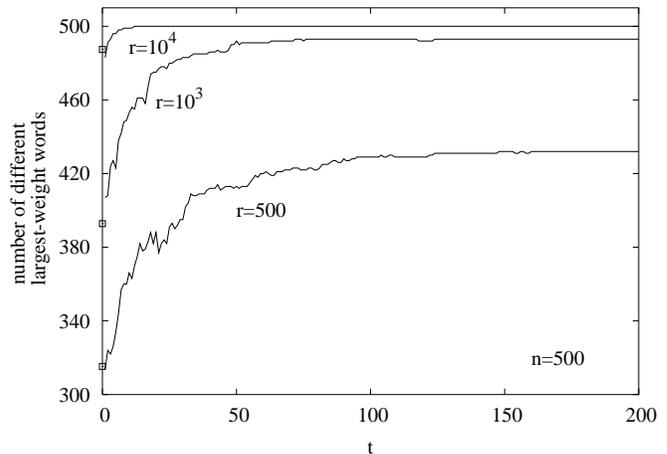} } \vspace{0cm} \caption{The time
evolution of the number of different largest-weight words for
$n=500$, $l=10$ and $\epsilon=10^{-5}$. Small squares at the $t=0$
axis indicate the values for the randomly drawn words (see text).
One can notice that during simulations a redistribution of
largest-weight words takes place and that reduces the number of
homonyms in the language. However, the number of largest-weight
words saturates below $n$ and that shows that homonyms are a
persistent feature of language. For large range $r$ the (almost)
homonymy-free language is obtained.} \label{timedifobj}
\end{figure}

Since homonymy typically occurs when more than one inventory has
the same largest-weight word, we examined in more detail the
number of different largest-weight words and the results are shown
in Fig.~\ref{timedifobj}. One can notice that as the interval $r$
from which the words are drawn increases, this number tends to the
number of objects $n$, and that means that homonyms become less
frequent. This is because for large $r$ there are many words to
chose from and the probability that two inventories have the same
largest-weight word decreases. 
Let us notice, however, that homonymy might appear also in some naming game models with an unbounded 
reservoir of words (in our case it corrsponds to the limit $r\rightarrow\infty$)~\cite{BRIGATTI}.

Naively, one might expect that the number of different largest-weight words can be obtained from the
simple probabilistic arguments: let us select randomly $n$ numbers
from the interval $<1,r>$ and check how many of them are
different. We did such calculations and numerical results are also
shown in Fig.~\ref{timedifobj} (small squares along the $t=0$
axis). One can notice that this agrees with simulations but only
initially. The subsequent evolution of the model changes the
initial distribution and the number of different largest-weight
words increases in time. Since success rate and the number of
different largest-weight words behave similarly
(Fig.~\ref{epsilon}a,b), such a redistribution reduces homonymy
and enables more efficient communication between agents. However,
saturation below the maximal value (seen in
Fig.~\ref{timedifobj}), equal to the number of objects $n$,
indicates that homonymy is a persistent feature of language.

\begin{figure}
\vspace{0.5cm} \centerline{ \epsfxsize=9cm\epsfbox{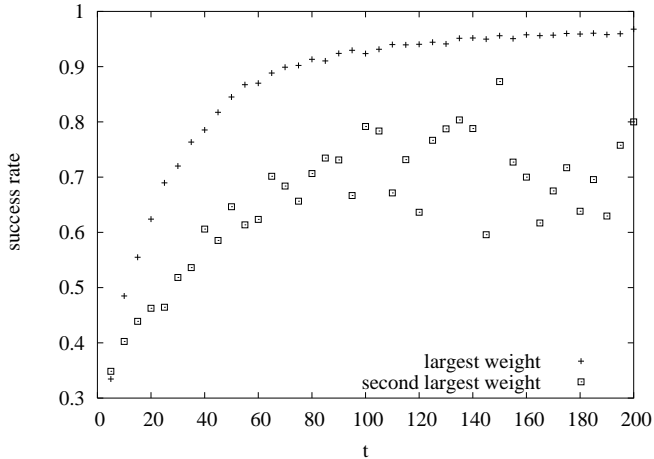}
} \vspace{0cm} \caption{The time evolution of the success rate of
utterances with largest- and second-largest-weight words.
Relatively large success rate of utterances with second-largest
weight words indicates that more than one word can be associated
with some objects, i.e., some words can be treated as synonyms.
Increasing in time fluctuations of the second-largest weight data
are due to poor statistics caused by the decreasing number of such
utterances (i.e, synonymy decreases over time). Simulations were made for $n=500$, $r=10^3$, $l=10$, and $\epsilon=10^{-5}$.}
\label{timel500z1000}
\end{figure}

Fig.~\ref{epsilon}c shows that a fraction of communication
attempts is made with the second-largest weight words. When such
an attempt is successful it usually means that there is more than
one word that  is associated with a given object, which for our
purposes defines synonymy. That such words do ensure a relatively
large success rate is confirmed in Fig.~\ref{timel500z1000}, where
the time evolution of the success rate of utterances with largest-
and second-largest-weight words is shown. Indeed, relatively large
success rate of utterances with second-largest weight words
indicates that more than one word can be associated with some
objects, i.e., some words can be treated as synonymous. However,
the decrease of frequency of second-largest-weight utterances seen
in Fig.~\ref{epsilon}c and (related with that) large fluctuations
seen in Fig.~\ref{timel500z1000} show that the role of synonyms
diminish in time. In the long-time limit synonymous
second-largest-weight words become irrelevant since entire
communication proceeds with largest-weight words only.

A trace of synonymy can be also seen in Fig.~\ref{config0}.
Indeed, overlapping plotting symbols (circles and crosses) show
that both agents have a substantial fraction of the same largest-
and second-largest-weight words in corresponding inventories. This
plot, however, does not tell us that for many of these pairs, the
weight of the largest-weight word is so much dominant that other
words from this inventory are essentially negligible (since they
are never used), especially after long simulations.
Although quantitative estimation of the role of homonymy and
synonymy depends on parameters, some generic behaviour seems to
characterize our model. In particular, homonymy, although rare for
large $r$, is a persistent  feature of the language: except for
the initial time interval, frequency of homonymous utterances
remains constant. On the other hand the frequency of synonymous
utterances decreases in time.

Provided that the model bears some similarity to the evolution of
natural languages,  one can expect that in present-day languages,
that correspond to the long-time limit of the language that emerge
in our model, synonymy, in agreement with some observations, would
be rare. It was already suggested by Hurford~\cite{HURFORD} that
rareness of synonymy is caused by the asymmetry of evolutionary
benefits between speaker and hearer. Let us emphasize that our
model uses only cultural (single-generation) mechanisms for the
evolution of language. The results thus show that understanding of
some basic features of homonymy and  synonymy can be obtained
within a much simpler model that does not take into account any
evolutionary effects.
Let us also notice that persistent homonymy and transient synonymy was also reported by Puglisi \emph{et al}.~\cite{PUGLISI} in an interesting model of category formation. In their model, that also includes cultural dynamics only, agents are exposed to the continuous environment (i.e., with an infinite number of objects) and such a behaviour was observed even when the reservoir of possible words is unbounded.
As a matter of fact transient nature of synonymy is a more generic property of the Naming Game, although in some models persistent synonymy was reported~\cite{BARON2007}.
\subsection{The effect of noise and distribution of words}
All calculations reported so far were made for the noiseless case
($p=0$), i.e., under the assumption that communication of a word
to another agent is perfect and cannot change the word. Now we
relax this assumption and examine the role of noise that might
distort the communicated word as specified in Eq.~(\ref{noise}).
In our opinion, especially at early stages of the evolution of
language communication could be exposed to such a disturbance.

\begin{figure}
\vspace{0cm} \centerline{ \epsfxsize=9cm \epsfbox{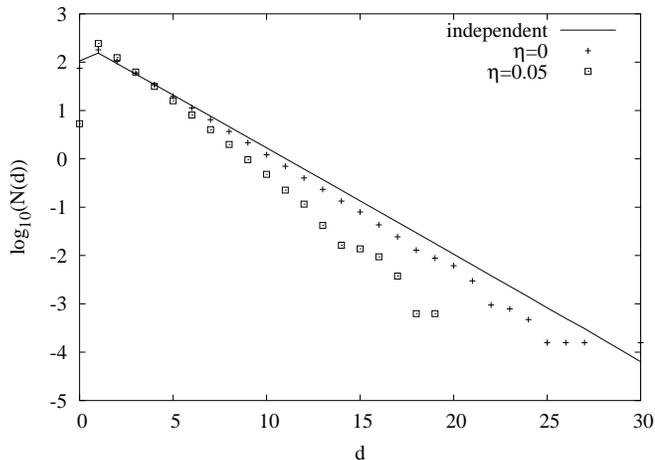} }
\vspace{0cm} \caption{The average number of intervals $N(d)$ of a
given distance $d$ between neighbouring largest-weight words.
Calculations were made for $n=500$ $r=10^3$, $l=10$, and
$\epsilon=10^{-5}$. This data show that noise greatly magnifies
redistribution of words so that they are more evenly distributed
within the available range (overlaps and large distances between
words are much less likely, comparing to the distribution where
they are selected independently).} \label{redistrib}
\end{figure}

Because of noise, the received word might be different than that
uttered by the speaker. If the difference is small, the hearer
might still correctly decode it. We expect that this will be often
the case when the amplitude of noise is small or the
largest-weight words are well separated so that the small change
does not lead to the overlap with some similar words. As we have
already noticed (Fig.~\ref{timedifobj}), during the evolution of
the model a redistribution of largest-weight words takes place,
that reduces homonymy and improves communication between agents.
Fig.~\ref{redistrib} shows that noise greatly magnifies such a
redistribution. In this figure we present the distribution of
distances $d$ between neighbouring largest-weight words compared
with the distribution where largest-weight words are selected
randomly.
 One can notice that noise leads to the more even distribution
 (within the available range) with
 substantially reduced number of overlaps ($d=0$) as well as of large voids.

\begin{figure}
\vspace{0.0cm} \centerline{ \hspace{0cm}\epsfxsize=9cm
\epsfbox{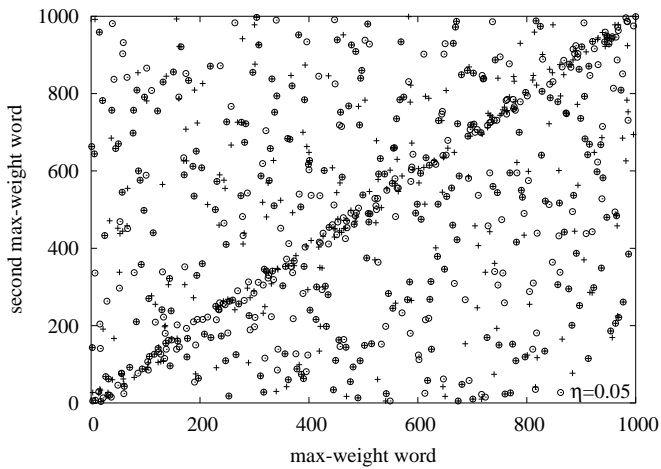} } \vspace{0cm} \caption{The distribution of
largest-weight words and the second-largest-weight words after
$t=10^3$ simulations. Calculations were made for $n=500$,
$r=10^3$, $l=10$, $\epsilon=10^{-5}$, and $\eta=0.05$. Different plotting symbols
correspond to different agents.} \label{config05}
\end{figure}

Noise also changes the distribution of second-largest-weight
words. Accumulation of points along the diagonal line seen in
Fig.~\ref{config05} shows that in presence of noise the second
largest-weight words are very often close to the largest-weight
words. In such a case they should not be considered as synonyms
(that are usually much different) but as the same words but e.g.,
with a slightly modified pronunciation. When noise is absent there
is no such accumulation (Fig.~\ref{config0}).

It is possible that noise played an important role in the
evolution of language and helped to redistribute words within
available phonetic space (Fig.~\ref{redistrib}) and/or reduced the
number of synonyms (Fig.~\ref{config05}). Actually, it would be
interesting to obtain the analogue of the distribution of
distances between words shown in Fig.~\ref{redistrib}, but
obtained for natural languages. Although the very definition of
distance between words remains under debate, various algorithms of
mainly  phonetic comparison are already in use~\cite{KESSLER} and
some statistical analysis in principle could be made.

\section{Conclusions}
In the present paper we studied an $n$-object naming game between
two agents and examined the structure of the emergent common
vocabulary. Our results show that after an initial transient a
linguistic synchronization is reached and efficient communication
of agents is established: speaker selects an object and a
corresponding word that is communicated to hearer that usually
correctly decodes the intended meaning. Our main results are
twofold:
\begin{itemize}
\item A small fraction of communication attempts use homonyms or
synonyms. Although homonyms reduce the efficiency of communication
they appear to be a rather persistent feature of the language. On
the other hand, synonyms do not reduce such an efficiency but are
gradually expelled from the language. The model supports thus the
observation that  nowadays in natural languages synonyms are rare, but related observations were also made in another type of models by Puglisi \emph{et al.}~\cite{PUGLISI}.
Moreover, it seems to us that the present model, that has only one
generation of agents and does not refer to the notion of fitness
is simpler than that used by Hurford~\cite{HURFORD} where the
rareness of synonymy was attributed to the asymmetry of payoff
between speaker and hearer.
 \item The second main result is to show that noise plays (or played)
 an important role in the evolution of language. It enhanced
 redistribution of words and probably contributed to the reduction
 of synonymy of the language.
\end{itemize}

It would be desirable to extend our model to a multi-agent. Let us notice, however, that such simulations
 are likely to be computationally very demanding (in such a case
the dynamics of the model will be slower and the amount of
 calculations needed for the model to reach the linguistic
 synchronization will be much larger). Additional problem might be related with examining the structure of emerging language and the present paper shows that even the preparatory (two-agent) version provides rich and nontrivial behaviour. Since, however, human linguistic interactions take place in multi-agent regime, such an extension should be examined. Another possibility might
 be to  introduce a fitness function and implement evolutionary
 changes that in some versions of naming game models are known to result in qualitatively
 novel behaviour~\cite{LIPLIP2008}. 
 Let us notice that our model neglects, among others, sound-merging effects as well as interactions of a given
 language with spatially neighbouring languages. Such factors often provide an important source of
 homonyms and synonyms. These factors might be taken into account in a multi-agent version of our
 model. Morever, it would be interesting to examine a situation where the number of objects $n$ could differ from the number of inventories, would depend on an agent and in addition would be determined in some dynamical process of category formation as in the paper of Puglisi \emph{et al.}~\cite{PUGLISI}.

\textbf{Acknowledgments:} This research was supported with Ministry of Science and Higher Education grant
N~N202~071435. We gratefully acknowledge access to the computing
facilities at Pozna\'n Supercomputing and Networking Center.

\end {document}